\documentclass[conference]{IEEEtran}
\ifCLASSINFOpdf
\else
\fi
\usepackage{amsmath}
\usepackage{amssymb,bm}
\usepackage{amsfonts}
\usepackage{algorithm}
\usepackage{algorithmicx}
\usepackage{algpseudocode}
\usepackage{multirow}
\usepackage{graphicx}
\usepackage{tabularx}
\usepackage{enumitem}

\begin{document}
%
\title{Audio Event and Scene Recognition: A Unified Approach using Strongly and Weakly Labeled Data}

\author{\IEEEauthorblockN{Anurag Kumar}
\IEEEauthorblockA{School of Computer Science\\
Carnegie Mellon University\\
Pittsburgh, PA-15213, USA\\
Email: alnu@andrew.cmu.edu}
\and
\IEEEauthorblockN{Bhiksha Raj}
\IEEEauthorblockA{School of Computer Science\\
Carnegie Mellon University\\
Pittsburgh, PA-15213,USA\\
Email: bhiksha@cs.cmu.edu}
}


%


\maketitle

\begin{abstract}
In this paper we propose a novel learning framework called \emph{Supervised and Weakly Supervised Learning} where the goal is to learn simultaneously from weakly and strongly labeled data. Strongly labeled data can be simply understood as fully supervised data where all labeled instances are available. In weakly supervised learning only data is weakly labeled which prevents one from directly applying supervised learning methods. Our proposed framework is motivated by the fact that a small amount of strongly labeled data can give considerable improvement over only weakly supervised learning. The primary problem domain focus of this paper is acoustic event and scene detection in audio recordings. We first propose a naive formulation for leveraging labeled data in both forms. We then propose a more general framework for \emph{Supervised and Weakly Supervised Learning} (SWSL). Based on this general framework, we propose a graph based approach for SWSL. Our main method is based on manifold regularization on graphs in which we show that the unified learning can be formulated as a constraint optimization problem which can be solved by iterative concave-convex procedure (CCCP). Our experiments show that our proposed framework can address several concerns of audio content analysis using weakly labeled data.
\end{abstract}


%
\IEEEpeerreviewmaketitle

\section{\textbf{Introduction}}
Capturing an audio or multimedia recording has become extremely easy and hence a huge amount of multimedia data is being generated every day. Audio either on its own or together with other modalities in a multimedia recording carry a significant amount of information. Hence, detection of acoustic events and scenes have become an important research problem \cite{stowell2015detection}. Several applications motivates audio event or scene detection research. These include but are not limited to content based indexing and retrieval of multimedia recordings \cite{schauble2012multimedia,yang2012multimedia,lew2006content}, improving human computer (robot) interaction by making computers (robots) aware of acoustic scene or environment around it \cite{janvier2012sound,maxime2014sound}, surveillance and monitoring application \cite{valenzise2007scream}, audio based context recognition system \cite{dargie2009adaptive}. 

Over the past few years, several approaches based on different signal processing and machine learning techniques have been proposed for audio event and scene detection, such as \cite{stowell2015detection,zhuang2010real,mesaros2015sound,phan2015random,gencoglu2014recognition} to cite a few. Audio event and scene detection challenges \cite{giannoulis2013detection} (DCASE 2013),  \cite{mesaros2016tut} (DCASE 2016) have helped in increasing the pace of audio content analysis research. 

However, almost all of current literature on audio event detection (AED) rely on fully supervised methods using \emph{strongly labeled}  data. In \emph{strongly labeled} data either audio clip examples for an acoustic event are directly provided or the time stamps of occurrences of the acoustic event in the recordings are given so that event specific part can be extracted from the whole recordings. Clearly, the purpose of strongly labeled  data is supervised training of event detectors. Overall, labeled examples of each event class are available and then some \emph{supervised} machine learning technique is applied for recognizing and detecting acoustic events. We will alternatively refer to \emph{strongly labeled} data as supervised data.  

This reliance on \emph{strongly labeled} data severely limits the scale and scope of audio event (and scene) detection works and is currently one of the most important challenges faced by the research community. Creating a large amount of strongly labeled data is an extremely time consuming, difficult and expensive process. This can be gauged from the fact that most publicly available datasets have less than an hour of audio data for each event \cite{piczak2015esc, mesaros2016tut,giannoulis2013detection, salamon2014dataset}. In fact, in most cases only a few minutes of audio data per acoustic event is available. Moreover, this also limits the vocabulary of audio events in datasets because of difficulties in creating strongly labeled data for a large number of acoustic event.

Recently, there have been attempts towards weakly supervised learning of audio events \cite{kumar2016audio}, \cite{anuragweakly}.  The goal is to learn audio event detectors from \emph{weakly labeled} data. In \emph{weakly labeled}  data only the presence or absence of an event in the recording is known. Exact time stamps of occurrences of events are not known and hence supervised learning methods cannot be directly applied. The methods proposed in these works are based on Multiple Instance Learning (MIL) framework. The motivation behind this weakly supervised learning is that not only manually annotating audio recordings for weak labels is much easier to do compared to strong labels but also that weakly labeled data can be directly obtained from web. Most of the multimedia or audio recordings on the web have some associated metadata (titles, tags etc) from which weak labels can be inferred to a certain extent. Thus, the time consuming and expensive process of manual annotation is not required any more and large amount of weakly labeled data can be directly obtained from web.   

In this work we propose a unified approach to learn simultaneously from weakly and strongly labeled data. Three important factors motivates this unified approach for learning audio event detectors. 

\begin{itemize}[leftmargin=*]
\item The problem with strongly labeled data is that it cannot be obtained or created in large amounts which further creates learning challenges. Nonetheless, in a lot of cases strongly labeled data are available or strong labeling can be done, though in a small amount. Weakly labeled data on the other hand can be obtained on a much larger scale and automatically from web. Given labeled data in two forms it is desirable to have a learning framework which can exploit them simultaneously. The labeled data in both forms can together help in learning robust models for audio events.  

\item The major significance of weakly supervised learning lies in the fact that it allows us to scale AED by providing a way to exploit the huge amount of data on web, example from \emph{Youtube}. Weak labels can be automatically obtained from the metadata associated with a recording. However, such weak labels are always expected to be noisy. Consider, for example, the sound event \emph{barking}. Searching \emph{barking} on Youtube also returns recordings such as \emph{Hillary Clinton literally barks at Republicans} among the top results which clearly has nothing to do with the acoustic event barking. Metadata based weak labels are always expected to be noisy. This label noise can adversely affect the learning process. SWSL can be one way to address these label noises. A small amount of  data can be strongly labeled where it is known that the labels are ``pure". It can be added to the pool of noisy weakly labeled data and this supervision can help learn better decision functions. The ill effects of noisy weak labels can be mitigated by the strongly labeled data or in other words presence of supervised data can make weakly supervised learning tolerant to label noises.   

\item  The multimedia data on web contains a large amount of within-category variations \cite{kumar2016audio}. Different recordings are recorded under different conditions and styles which produces huge intra-class variations. Recording instances of the same event vary by a large amount. This makes the learning process from such data very challenging. Moreover, the audio signal itself might be very noisy. Overall, learning from web data introduces yet another challenge in form of what we can refer to as ``signal noise".  It includes the large amount of intra-class variations for a given sound event class.  Once again, a small amount of ``pure" examples from strongly labeled data can be an answer to these concerns. 

\end{itemize}

\noindent Hence, a novel unified learning framework which can leverage labeled data in both forms is proposed in this paper. We call this novel unified learning framework \emph{Supervised and Weakly Supervised Learning} (SWSL). Under this framework, we propose two leaning methods. The first one called \emph{simpleSWSL} or \emph{naiveSWSL} is a naive way to bring supervised data into weakly supervised domain. It adapts the supervised data in MIL framework to learn simultaneously from both forms of labeled data. We then give a more general learning process for SWSL. The central idea behind this second approach is that SWSL can be formulated as semi-supervised learning with constraints imposed by weak labels. Under this formulation, one can adapt a variety of semi-supervised learning for SWSL. In particular, we use graph based semi-supervised learning and refer to this method as \emph{graphSWSL}. 

It should be noted that although this paper is driven by audio event detection, our \emph{SWSL} approach is a completely generic framework to learn using both fully supervised and weakly labeled data. It can be applied to other related problems as well. For example, multiple instance learning based weakly supervised image retrieval and visual object detection  have been explored in computer vision community \cite{cinbis2016weakly}, \cite{vijayanarasimhan2008keywords,zhang2002content}. Our SWSL framework can be applied in these domains as well. 

The outline of the paper is as follows; in Section 2 we introduce the proposed SWSL framework. In Section 3 we describe graphSWSL formulation. Section 4 describes the audio feature representation used in this work. Section 5 shows and discusses our experiments and results. Finally we conclude in Section 6. 

\section{\textbf{SWSL}}
\cite{kumar2016audio} introduced audio event detection (AED) using weakly labeled data. The main idea in weakly supervised AED is that AED using weak labels can be formulated as an multiple instance learning problem \cite{andrews2002support}. In MIL instances are given in groups called \emph{bags} and labels are available for each bag. In a negative bag all instances are known to be negative, whereas, in a positive bag it is only known that at least one instance is positive. Thus, in a positive bag both positive and negative instances can be present. The goal is to learn a classifier technique using data in bag-label form. 

For AED, full audio recordings are segmented into small segments. The full recording is a bag and the segments of the recording are instances of the bag. Since weak label carries information about presence or absence of an event in a recording, it can be used to label bags (recordings). For an event, a bag is positive if that event is marked to be present in the recording otherwise it is labeled as a negative bag. \cite{kumar2016audio} used SVM (miSVM) \cite{andrews2002support} and neural network (BPMIL) \cite{zhou2002neural} based MIL methods to show that AED with weak labels can be successfully done. Besides learning from weak labels, another major advantage of these methods is that they can be used to estimate temporal location of events as well. The significance of it is due to the fact that no such information was present in the weakly labeled data to begin with. In another work, \cite{anuragweakly} tries to scale AED with weak labels by proposing scalable MIL methods. These scalable MIL methods convert MIL into supervised learning by representing each bag with a single high dimensional vector. Although, a significant reduction in computational time is achieved, these scalable methods operate at bag-level and cannot be used for temporal localization of events. 

For SWSL, we retain the basic MIL based framework to work with weakly labeled data. Audio recordings are converted into bags by segmenting into small segments and positive (+1) or negative (-1) labels are assigned to the bags according to weak labels. In SWSL, along with weak data in form of labeled bags, we are also given a separate strongly labeled or supervised data in form of labeled instances. The idea is to be able to exploit both supervised and weakly supervised data for AED.
\subsection{Naive SWSL}
In SWSL, along with weak data in form of labeled bags, we are also given a separate strongly labeled or supervised data in form of labeled instances. The simplest approach for SWSL is to formulate the strongly labeled data as a special case of weakly labeled data. In this special case, each labeled instance is framed as a bag with one instance only. The bag label is same as instance level. Once this is done, any MIL approach can be applied as in case of only weakly supervised learning. We call this method \emph{naiveSWSL} or \emph{simpleSWSL}.

Since each bag contain only one instance, all MIL methods factoring in bag constraints of at least one positive in positive bag and all negatives in negative bag will end up satisfying supervised label constraints. For example, using miSVM as simpleSWSL would imply that for the bags formed from supervised data, label constraints as in classical supervised SVM are satisfied whereas for other bags it would remain same as in miSVM. Hence, data in both strong and weak forms are appropriately used under this formulation. 

\subsection{Generalized SWSL}
The above formulation is a very naive way of unifying supervised and weakly supervised learning. In this section, we present a more general framework for SWSL. In MIL negative bags are known to have only negative instances, meaning labels for all instances in negative bags are essentially known. Thus, instances in negative bags can be considered along with supervised data. 

For positive bags on the other hand this is not valid. In generalized SWSL formulation, we undertake instances in positive bags as essentially unlabeled but with certain ``label constraints". The constraint on \emph{this unlabeled data} is that they are grouped into bags and within each bag of instances at least one instance is positive. Thus, we have labeled data in form of supervised data plus instances in negative bags and a set of unlabeled data with constraints. Hence, this general form of SWSL can be formulated  under the framework of semi-supervised learning with additional label constraints on the \emph{unlabeled} data. A variety of methods for semi-supervised learning including multiple instance learning semi-supervised learning have been proposed \cite{zhu2009introduction,zhu2003semi,belkin2006manifold,bennett1999semi,rahmani2006missl,zhou2007relation,jia2008instance}. In this work, we adopt one of the most popular method for semi-supervised learning, manifold regularization on graphs \cite{belkin2006manifold} for SWSL. We name this variant of SWSL as \emph{graphSWSL}.

\section{\textbf{graphSWSL: Graph Based SWSL}}

All instances in negative bags can be labeled as negative and hence from here on in our mathematical formulation we will simply consider them to be part of strongly labeled or fully supervised set.  

Let us represent the supervised dataset as $\mathcal{D}_s=\{(\mathbf{x}_1,y_1),...(\mathbf{x}_n,y_n)\}$. $y_i \in \{-1,1\}$ is label for instance $\mathbf{x}_i$ and $n$ is total number of instances in $\mathcal{D}_s$. The weakly supervised data $\mathcal{D}_w$, is in form of positive bags. Let $\mathcal{D}_w = \{B_1,..B_T\}$ be the set of $T$ bags where $B_t=\{(\mathbf{x}_{t1},y_{t1}),...,(\mathbf{x}_{tm_t},y_{tm_t})\}$ is a positive bag of instances where labels $y$ are unknown but at least one label is $+1$. $m=\sum_{t=1}^T m_t$ is the total number of instances in all positive bags. 

Let us represent the over all data $\mathcal{D}$ as $\mathcal{D}=\{(\mathbf{x}_1,y_1),...(\mathbf{x}_n,y_n),(\mathbf{x}_{n+1},y_{n+1}),....(\mathbf{x}_{n+m},y_{n+m})\}$. Without loss of generality we have assumed that instances are ordered such that first $n$ are from $\mathcal{D}_s$ and $n+1$ to $n+m$ are from $\mathcal{D}_w$. Instances $n+1$ to $n+m_1$ are from bag $B_1$ and so on. We will denote the start and end indices of instances from bag $B_t$ in $\mathcal{D}$ as $p_t$ and $q_t$. The instance space is denoted by $\mathcal{X}$. The total number of instances in $\mathcal{D}$ is $N=n+m$.

Labels for the first $n$ instances in $\mathcal{D}$ are known. The labels for the rest of the instances are unknown but constraint by the following relationship
\begin{equation}
\label{eq:bgcon}
max(y_{p_t},...y_{q_t}) = 1 \,\,\, \forall \,\, t=1 \,\,to\,\,T
\end{equation}

In this supervised and weakly supervised paradigm the goal is to learn the function mapping $\mathit{f}:\mathcal{X} \rightarrow R$, which maps the instance space to a decision score. $\mathit{f}$ is assumed to be smooth and let us denote the Reproducing Kernel Hilbert Space (RKHS) of $\mathit{f}$ as $\mathcal{H}$. 

Since the labels for instances $\mathbf{x}_{n+1}$ to $\mathbf{x}_{n+m}$ are essentially unknown and yet constrained by Eq \ref{eq:bgcon},  we can formulate this learning process as a constrained form of semi-supervised learning (SSL). A particularly well known method for semi-supervised learning is manifold regularization on graphs \cite{belkin2006manifold}. This method is inductive which is one of the reasons we adopt it for SWSL. 

\subsection{Manifold Regularization approach for SWSL} 

In Graph based semi-supervised learning, all instances are assumed to connected by a graph $G=(V,E)$ where the vertices $V$ are instances in the data. In this paper we assume kNN graph \cite{zhu2009introduction} where a vertex $\mathbf{x}_i$ is connected to another vertex $\mathbf{x}_j$ by a non-zero weight $w_{ij}$ if $\mathbf{x}_i$ is among the $k$-nearest neighbour of $\mathbf{x}_j$ and vice versa. 

The edge weight $w_{ij}$ is then defined by Gaussian Kernel, $w_{ij} = exp(-\frac{||\mathbf{x}_i - \mathbf{x}_j||^2}{2\sigma^2})$. $\sigma$ is the bandwidth parameter for weights. Clearly, when $\mathbf{x}_i$ and $\mathbf{x}_j$ are not connected $w_{ij}=0$. The overall graph is parametrized through a symmetric weight matrix $W$ whose elements are $w_{ij}$. Finally, the unnormalized graph laplacian $L$ is defined by $L = D - W$, where $D$ is diagonal matrix with $D_{ii} = \sum_j w_{ij}$. 

Manifold regularization on graphs for SSL solves the following optimization problem
\begin{equation}
\underset{f}{min} \,\,\, \frac{1}{n} \sum_{i=1}^n (y_i - \mathit{f}(\mathbf{x}_i))^2 + \lambda_1 ||\mathit{f}||_{\mathcal{H}}^2 + \lambda_2 ||\mathit{f}||_{I}^2 
\end{equation}
The first term is simply the squares loss over the labeled instances. The first regularization term $||\mathit{f}||_H^2$ is the RKHS norm which is used to impose smoothness conditions on $\mathit{f}$. The second penalty term $||\mathit{f}||_{I}^2$ is a regularization term for intrinsic structure of data distribution. This terms ensures that the solution is smooth with respect to data distribution as well.  Together the two regularization terms controls complexity of the solution over both RKHS and intrinsic geometry of data distribution. 

For SWSL, we need to factor in the weak label information of positive bags in the above optimization problem. To do this, we solve following optimization problem in manifold regularization on graphs for SWSL
\begin{align}
\label{eq:mainopt}
\underset{f}{min} \,\,\, & \frac{1}{n} \sum_{i=1}^n (y_i - \mathit{f}(\mathbf{x}_i))^2 + \lambda_1 ||\mathit{f}||_{\mathcal{H}}^2 + \lambda_2 ||\mathit{f}||_{I}^2 \nonumber \\
& + \frac{\lambda_3}{T}\sum_{t=1}^T (1- \underset{j=p_t,...,q_t}{max} f(x_j))^2 
\end{align}
In the above formulation, the last term is the squared loss for each positive (+1) bag and factors in the weak label information. To measure loss with respect to each bag, the value of each bag is determined by the maximal output instance.  

Unlike the case of semi-supervised learning, the above SWSL formulation is a non-convex optimization problem. This presents additional challenges and here we describe an approach to solve the above optimization problem. 

First, we rewrite the optimization problem in Eq \ref{eq:mainopt} using slack variables. 
\begin{align}
\label{eq:reopt}
\underset{f,\mathbf{\xi}}{min} \,\,\, \sum_{i=1}^n & (y_i - \mathit{f}(\mathbf{x}_i))^2 +  \lambda_1 ||\mathit{f}||_{\mathcal{H}}^2 + \lambda_2 ||\mathit{f}||_{I}^2  + \lambda_3\sum_{t=1}^T \xi_t^2 \nonumber\\
s.t \,\, & \,\, 1 - \underset{j=p_t,...,q_t}{max} f(x_j) \leq \xi_t, \,\,\, t=1,...,T \\
& \,\, \xi_t \geq 0, \,\,  t=1,...,T \nonumber 
\end{align}

$\xi_t$ are the slack variables for loss on positive bags. Also, note that we have factored in the normalization terms ($n$ and $T$) in the regularization parameters. To solve the above problem we need a finite dimensional form for $f$. 

Using Representer Theorem \cite{smola1998learning}, the solution to the above problem can be expressed as $f(\mathbf{x}) = \sum\limits_{i=1}^{N} \alpha_i k(\mathbf{x},\mathbf{x}_i)$, where $k(\cdot,\cdot)$ is the reproducing kernel of $\mathcal{H}$. Let us denote the $N \times N$ kernel gram matrix over the training data $\mathcal{D}$ with $K$. 

Now, let $Y$ be a $N$ dimensional label vector where $Y = [y_1,y_2,...y_n,0,0....0]$. $Y_i$ is label for the first $n$ instances which are labeled and $0$ for the rest. Also, let $J$ be $N \times N$ diagonal matrix where $J_{ii}=1$ for $i=1\,\,to\,\,n$. and $J_{ii}=0$ for $i=n+1\,\,to\,\,(n+m)$. Using expression for $f$, the squared loss term for labeled instances can be written as $\sum_{i=1}^n  (y_i - \mathit{f}(\mathbf{x}_i))^2 = (Y-JK\alpha)^T(Y-JK\alpha)$. 

The intrinsic norm $||\mathit{f}||_I$ is estimated using the graph laplacian matrix $L$ by $||\mathit{f}||_I^2 = \frac{1}{N^2} f^TLf$ \cite{belkin2006manifold}. Hence, $||\mathit{f}||_I^2 = \frac{1}{N^2} \alpha^TKLK\alpha$. So our, final optimization problem becomes
\begin{align}
\label{eq:finopt}
\underset{\alpha,\mathbf{\xi}}{min} \,\,\, & (Y-JK\alpha)^T(Y-JK\alpha) +  \lambda_1 \alpha^TK\alpha \nonumber \\
& + \lambda_2 \frac{1}{N^2} \alpha^TKLK\alpha  + \lambda_3\sum_{t=1}^T \xi_t^2 \\
s.t \,\, & \,\, 1 - \underset{j=p_t,...,q_t}{max} K_j'\alpha \leq \xi_t \,\,\,\, t=1,...,T \nonumber \\ 
\,\, & \xi_t \geq 0, \,\, t=1,...,T \nonumber
\end{align}
$K_j$ is the $j^{th}$ column of kernel matrix $K$. The optimization problem in Eq \ref{eq:finopt} is still not straightforward to solve due the \emph{max} constraints. 
\vspace{-0.05in}
\subsection{Optimization Solution}
The objective function in the optimization problem in Eq \ref{eq:finopt} is a convex differentiable function. The first set of constraint ($1 - \underset{j=p_t,...,q_t}{max} K_j'\alpha \leq \xi_t$) is non-convex but  a difference of two convex function. Convex Concave Procedure (CCCP) \cite{smola2005kernel} is a well known method of sequential convex programming to handle problems like this. It is an iterative method in which the non-convex function is converted into a convex function using Taylor series approximation at the current solution. 

For an objective or constraint in form of $g(x) - h(x)$ where $g(x)$ and $h(x)$ are convex, a convex approximation at $x^{(k)}$ is obtained as $g(x) - h(x^{(k)}) - \triangledown h(x^{(k)}) (x - x^{(k)}) $, where $\triangledown h(x^{(k)})$ is gradient of $h(x)$ at $x^{(k)}$.  $max()$ is a non-smooth function and hence we need the subgradient of $max()$ for the Taylor series expansion. The subgradient of $\underset{j=p_t...q_t}{max} K_j'\alpha$ can be defined as \cite{cheung2006regularization}
\begin{equation}
\partial(\underset{j=p_t,...,q_t}{max} K_j'\alpha) = \sum\limits_{j=p_t}^{q_t} \delta_{tj}K_j 
\end{equation}
The $\mathbf{\delta}_{tj}$'s are defined as
\begin{equation}
\delta_{tj} = 
\begin{cases}
\frac{1}{r_t}, & \text{if } K_j'\alpha = \underset{u=p_t,...,q_t}{max} K_u'\alpha \\
0,  & \text{otherwise}
\end{cases}
\end{equation} 
$r_t$ is the number of instances which maximizes the output $K_j'\alpha$ in $t^{th}$ bag. Hence, all instances in bag for which maximum is achieved are active in the subgradients. Now, we can rewrite the non-convex constraints in $k^{th}$ iteration of CCCP using the above subgradient. The non-convex constraint $1 - \underset{j=p_t,...,q_t}{max} K_j'\alpha $ in $k^{th}$ iteration becomes
\begin{equation}
\resizebox{0.9\columnwidth}{!}{$1 - \underset{j=p_t...q_t}{max} K_j'\alpha \approx 1 - (\underset{j=p_t...q_t}{max} K_j'\alpha^{(k)} + \sum\limits_{j=p_t}^{q_t} \delta_{tj}^{(k)}K_j'(\alpha - \alpha^{(k)}))$}
\end{equation}

Hence, the final optimization problem to solve in $k^{th}$ iteration of CCCP
\begin{align}
\label{eq:cccpopt}
\underset{\alpha,\mathbf{\xi}}{min} \,\,\, & (Y-JK\alpha)^T(Y-JK\alpha) +  \lambda_1 \alpha^TK\alpha \nonumber \\
& + \lambda_2 \frac{1}{N^2} \alpha^TKLK\alpha  + \lambda_3\sum_{t=1}^T \xi_t^2 \\
& s.t \nonumber \\
&  1- (\underset{j=p_t...q_t}{max} K_j'\alpha^{(k)} + \sum\limits_{j=p_t}^{q_t} \delta_{tj}^{(k)}K_j'(\alpha - \alpha^{(k)})) \leq \xi_t \nonumber \\ 
		&  \,\, \hspace{2.0in}  t=1,...,T \nonumber \\
\,\, & \xi_t \geq 0, \,\, t=1,...,T \nonumber
\end{align}
The objective function in optimization problem of Eq \ref{eq:cccpopt} is convex and the constraints are linear. The overall problem is a convex Quadratic Programming problem. In CCCP the above optimization problem is iteratively solved until convergence. 

Once $\alpha$ has been obtained the output corresponding to any test point $\mathbf{x}_{test}$ can be obtained as $f(\mathbf{x}_{test}) = \sum_{i=1}^N \alpha_i k(\mathbf{x}_{test},\mathbf{x}_i)$. It is worth noting that similar to miSVM and neural network based weakly supervised approaches used for AED, graphSWSL can also predict output corresponding to each instance in a bag. Hence, our SWSL approach can also be used for temporal localization of acoustic events in a recording. The bag-level prediction is done by using the $max$ over instance outputs. 
\section{Acoustic Features for Audio Segments}
\label{sec:acofeat}
We need a feature representation for each audio segment or instances in bags. Several acoustic feature representations for audio event or scene detection has been proposed \cite{stowell2015detection}. In this work we use the Gaussian Mixture Model (GMM) based histogram characterization of audio segments which has been used in previous weakly supervised audio event detection works \cite{anuragweakly},\cite{kumar2016audio}. It has been shown to be effective feature representation for audio event detection tasks. Components of GMM represent \emph{audio words} and the final features are normalized soft-count histograms. 

All audio recordings are first parameterized through Mel-Ceptra Coefficients (MFCCs). The MFCCs for the training data are then used to train a GMM. Let $\mathcal{G} = \{w_c,N(\vec{\mu}_c,\Sigma_c)\}$ be this GMM, where $w_c$ is the mixture weight for $c^{th}$ component of GMM and $\vec{\mu}_c$ are $\Sigma_c$ are the component mean and covariance matrix. Let the total number of components in GMM be $C$. We  train GMM with diagonal covariance matrices $\Sigma_c$. Now, given an audio segment with a total of $M$ MFCC frames where each frame is represented by $\vec{f}_t$, we compute the following for each GMM component
\begin{align}
Pr(c | \vec{f_{t}}) = & \frac{w_{c}N(\vec{f_{t}} ; \vec{\mu}_c, \Sigma_c)}{\sum\limits_{i=1}^C w_i N(\vec{f_{t}} ; \vec{\mu}_i, \Sigma_i)}\\
H(c) = & \frac{1}{M}\sum\limits_{t=1}^M Pr(c | \vec{f_{t}})
\end{align}
The final feature for the audio segment is the  $C$ dimensional histogram vector $\vec{H}_C=[H(1),...,H(C)]^T$. It is also usually helpful to normalize  $\vec{H}_C$ to sum to 1.

\section{Experiments and Results}

We evaluate the proposed supervised and weakly supervised learning on both audio event and acoustic scene detection tasks. In our experiments we add a small amount of strongly labeled data to the pool of weakly labeled data to learn event or scene detectors using SWSL and compare it with weakly supervised learning. For weakly supervised learning we use miSVM approach \cite{andrews2002support}. miSVM has also been used in previous weakly supervised audio event detection work \cite{kumar2016audio}. For simpleSWSL, we again use miSVM approach to integrate strongly and weakly labeled data.  

The details specific to audio event and scenes are given in corresponding sections. We describe the common experimental set up here. In our experiments all audio recordings are sampled at $44.1KHz$ sampling frequency. $20$ dimensional MFCC features along with delta and acceleration coeffecients are used to parametrize audio recordings. The bag of audio words over the MFCCs as described in Section \ref{sec:acofeat} are used as feature representation for audio segments or instances in bags. The GMM component size $C$ is $64$ or $128$. We show results for both values of $C$. 

Exponential Chi-square ($\chi^2$) kernels in form of $exp(-\gamma d(x,y))$ have been known to work remarkably well with histogram features, including for detection of acoustic concepts \cite{rawat2013robust}\cite{kumar2016features}. $d(x,y)$ is $\chi^2$ distance. Hence, we use exponential $\chi^2$ kernels for miSVM and simpleSWSL. The parameter $\gamma$ is set as the inverse of mean $\chi^2$ distance between training points. The slack parameter $C$ in SVM training is set through cross validation on the training data. 

For graphSWSL, the kNN graph is constructed with $k$ as $20$ or $40$. We evaluate and show results for both cases. The bandwidth $\sigma$ for graph weights is set to $1.0$ for all experiments. The kernel $K$ is again exponential $\chi^2$ kernel as used in miSVM and simpleSWSL. The parameter $\lambda_3$ in graphSWSL is simply set as $\lambda_3 = n/T$, ratio of the number of supervised instances to the number of positive bags. The other two regularization parameters $\lambda_1$ and $\lambda_2$ are selected through cross-validation over a grid of $10^{-3}$ to $10^3$. 
\subsection{Acoustic Event Detection}
\label{sec:aedexpt}

We consider a set of $10$ acoustic events namely, \emph{Chainsaw (C), Clock Ticking (CT), Crackling Fire (CF), Crying Baby (CR), Dog Barking (DB), Helicopter (H), Rain (RA), Rooster (RO), Seawaves (SE), Sneezing (SN)}. The events are part of ESC-10 \cite{piczak2015esc} dataset which provides us strongly labeled data for these events. We obtain the weakly labeled training data from Youtube. For each event we use the event name as search query on Youtube \footnote{www.youtube.com}. To get more sound oriented results the keyword ``sound" is attached to each event name (e.g Chaisaw sound). We consider the audio from top $60$ returned video results, from which we filter out very long recordings. Finally, we are left with $40$ weakly labeled recording for all acoustic events except \emph{Crackling Fire} (35) and \emph{Rain}(10). Recordings for Rain are relatively longer and hence total duration of audio for it is in similar range to others. The total duration of all $365$ recordings is around $5.1$ hours. Audio recording are segmented to form bags and instances and we use the average duration of each event in the strongly labeled data as the segment size. 

We need a large test dataset for comprehensive evaluation of all methods. For all audio events, we obtain test data from \emph{Freesound} \footnote{www.freesound.org}.  The number of test recordings for each event are as follows: C (78), CT (69), CF (66), CR (88), DB (100), H (39), RA (76), RO (83), SE (48), SN (75). This total of $722$ \emph{recordings} spanning over $13$ \emph{hours} of audio allows a thorough analysis of all methods. Note that results given in next paragraphs are bag-level detection results. Instance level temporal localization experiments are described in further sections.

ESC-10 dataset contains $40$ positive examples for each event, resulting in a total of $400$ supervised data instances (positive and negative) for any given event. The dataset comes pre-divided into $5$ folds. We include recordings from $4$ out of $5$ folds for training. This amounts to addition of about $25$ minutes of strongly labeled data to the pool of weakly labeled data. Experiments are run all $5$ ways (leaving one fold in each case) and the average across all $5$ runs are reported. We use Average Precision (AP) as performance metric. The mean average precision (MAP) over all event classes are also shown for all cases. 
 
\begin{table}[t]
\centering
\caption{Results (AP) using miSVM and simpleSWSL}
\label{tab:aemswsl}
\resizebox*{1.0\columnwidth}{!}{
\begin{tabular}{|c|c|c|c|c|}
\hline  
Events &\multicolumn{2}{c|}{$C=64$} & \multicolumn{2}{c|}{$C=128$}\\ 
\cline{2-5}
 & miSVM & simpleSWSL & miSVM & simpleSWSL\\
\hline
Chainsaw&0.571&0.671&0.436&0.649\\
\hline
Clock Ticking&0.563&0.658&0.542&0.689\\
\hline
Crackling Fire&0.382&0.458&0.421&0.522\\
\hline
Crying Baby&0.558&0.630&0.613&0.691\\
\hline
Dog Barking&0.237&0.348&0.442&0.520\\
\hline
Helicopter&0.363&0.384&0.393&0.431\\
\hline
Rain&0.263&0.414&0.252&0.374\\
\hline
Rooster&0.392&0.444&0.466&0.533\\
\hline
Seawaves&0.164&0.162&0.176&0.171\\
\hline
Sneezing&0.320&0.402&0.327&0.424\\
\hline
\textbf{MAP}&\textbf{0.381}&\textbf{0.457}&\textbf{0.407}&\textbf{0.500}\\
\hline
\end{tabular}
}
\end{table}
Table \ref{tab:aemswsl} shows AP values for different audio events using miSVM and simpleSWSL.  Results for acoustic features using both $C=64$ and $C=128$  are shown. One can observe that adding a small amount of strongly labeled data to the pool of weakly labeled data is extremely helpful. An absolute improvement in MAP of about $\mathbf{7\%}$ for $C=64$ and $\mathbf{10\%}$ for $C=128$ can be observed. As far as individual events are concerned, for several events such as \emph{Chainsaw, Crackling Fire, Rain} an absolute improvement of $15-20\%$ in AP can be observed.  

Table \ref{tab:aegswsl} shows AP values for graph based SWSL approach. We observe that graphSWSL improves over simpleSWSL another $4-5\%$ in absolute terms. This amounts to about $\mathbf{12\%}$ and $\mathbf{14\%}$ improvement over miSVM for $C=64$ and $128$ and respectively. For graphSWSL the performance remains more or less consistent for the two values of $k$ in $kNN$ graph. 

From these results one can conclude that a small amount of strongly labeled can play a significant role in reducing the effect of signal-noise and label-noise in weakly labeled data obtained from web. 
\begin{table}[t]
\centering
\caption{Results (AP) using graphSWSL}
\label{tab:aegswsl}
\resizebox*{1.0\columnwidth}{!}{
\begin{tabular}{|c|c|c|c|c|}
\hline  
Events &\multicolumn{2}{c|}{$C=64$} & \multicolumn{2}{c|}{$C=128$}\\ 
\cline{2-5}
 & $kNN=20$& $kNN=40$ & $kNN=20$ & $kNN=40$\\
\hline
Chainsaw & 0.534&0.531&0.578&0.574\\
\hline
Clock Ticking & 0.669&0.672&0.713&0.713\\
\hline
Crackling Fire & 0.571&0.584&0.579&0.574\\
\hline
Crying Baby & 0.749&0.741&0.767&0.772\\
\hline
Dog Barking & 0.305&0.305&0.439&0.439\\
\hline
Helicopter & 0.448&0.458&0.565&0.526\\
\hline
Rain & 0.421&0.403&0.382&0.414\\
\hline
Rooster & 0.612&0.610&0.695&0.678\\
\hline
Seawaves & 0.178&0.174&0.194&0.198\\
\hline
Sneezing & 0.523&0.523&0.519&0.519\\
\hline
\textbf{MAP} & \textbf{0.501}&\textbf{0.500}&\textbf{0.543}&\textbf{0.541}\\
\hline
\end{tabular}
}
\end{table}

\subsection{Acoustic Scene Detection}
The procedure for acoustic scene remains similar to acoustic events. We work with a toal of $15$ acoustic scenes from DCASE \cite{mesaros2016tut} dataset, which is also source of our strongly labeled data. Weakly labeled training data is again obtained from Youtube in a procedure similar to audio events. In this case we create weakly labeled data of $40$ recordings per scene, totaling $600$ recordings which spans over $27$ hours. Once again we obtain test data from Freesound. A total of $928$ test recordings ($38$ hours) are used. The test data contains an average of $61$ recording per scene with a minimum of $53$ for \emph{Forest Path} scene and a maximum of $71$ for \emph{Residential Area} scene. Segment size is average duration of scenes in strongly labeled data. The supervised data from DCASE comes pre-divided into $4$ folds and our experimental approach is same as before. We use $3$ out of $4$ folds in SWSL and perform experiments all 4 ways. As before, average results across all $4$ runs are reported here. The value of $k$ in $kNN$ graph is $40$. 

It is worth noting that acoustic scenes are acoustically much more complex. Intra-class variation is far greater than audio events. Both training and test data contains a significant amount of within class variations and hence, learning and detection of acoustic scenes from weakly labeled data is a much harder problem. This is evident in the low average precision for most classes. 

Table \ref{tab:ascene} shows AP values for different methods. For acoustic scenes, we note that the performance of miSVM and simpleSWSL is almost similar in terms of mean average precision. Graph based SWSL on the other hand gives a $\mathbf{20\%}$ relative improvement over these methods. graphSWSL improves AP in almost all cases. For several scenes such as \emph{Grocery Store, Home, Library, Metro Station, Train, Tram} AP improves by $\mathbf{50\%}$ or more in relative terms. This should be especially noted for \emph{Home}, which turns out to be the hardest scene to detect. graphSWSL for this scene more than \emph{double} the average precision when compared to weakly supervised learning. Again, based on these results we can conclude that SWSL is a very effective way of addressing the concerns of weakly supervised learning.  
\begin{table}[t]
\centering
\caption{Acoustic Scene Results (AP). graphSWSL (gSWSL), simpleSWSL (sSWSL)}
\label{tab:ascene}
\resizebox*{1.0\columnwidth}{!}{
\begin{tabular}{|c|c|c|c|c|c|c|}
\hline  
Events &\multicolumn{3}{c|}{$C=64$} & \multicolumn{3}{c|}{$C=128$}\\ 
\cline{2-7}
 & miSVM & sSWSL & gSWSL & miSVM & sWSL & gSWSL\\
\hline
Beach & 0.129&0.132&0.153&0.140&0.150&0.122\\
\hline
Bus & 0.087&0.099&0.110&0.094&0.104&0.106\\
\hline
Cafe & 0.264&0.220&0.256&0.272&0.246&0.301\\
\hline
Car & 0.066&0.083&0.078&0.069&0.085&0.079\\
\hline
City Center & 0.137&0.122&0.121&0.132&0.129&0.121\\
\hline
Forest & 0.074&0.076&0.081&0.090&0.095&0.092\\
\hline
Grocery Store & 0.053&0.061&0.081&0.059&0.066&0.098\\
\hline
Home & 0.048&0.060&0.116&0.049&0.067&0.098\\
\hline
Library & 0.073&0.106&0.113&0.070&0.086&0.088\\
\hline
Metro Station & 0.090&0.096&0.133&0.086&0.096&0.119\\
\hline
Office & 0.105&0.093&0.080&0.099&0.091&0.069\\
\hline
Park & 0.115&0.119&0.133&0.129&0.153&0.142\\
\hline
Residential Area & 0.092&0.112&0.099&0.085&0.113&0.111\\
\hline
Train & 0.070&0.066&0.105&0.074&0.068&0.092\\
\hline
Tram & 0.106&0.132&0.144&0.111&0.144&0.157\\
\hline
\textbf{MAP} & \textbf{0.101}&\textbf{0.105}&\textbf{0.120}&\textbf{0.104}&\textbf{0.113}&\textbf{0.120}\\
\hline
\end{tabular}
}
\end{table}
\subsection{Temporal Localization of Audio Events}
\begin{table}[t]
\centering
\caption{Temporal Localization Results}
\label{tab:temploc}
\resizebox*{1.0\columnwidth}{!}{
\begin{tabular}{|c|c|c|c|c|c|c|}
\hline  
Events &\multicolumn{3}{c|}{$C=64$} & \multicolumn{3}{c|}{$C=128$}\\ 
\cline{2-7}
 & miSVM & sSWSL & gSWSL & miSVM & sWSL & gSWSL\\
\hline
Chainsaw&0.455&0.646&0.497&0.372&0.640&0.668\\
\hline
Clock Ticking & 0.702&0.766&0.846&0.704&0.856&0.800\\
\hline
Cracking Fire & 0.715&0.841&0.914&0.691&0.878&0.886\\
\hline
Crying Baby & 0.861&0.923&0.980&0.846&0.931&0.983\\
\hline
Dog Barking & 0.510&0.726&0.772&0.621&0.810&0.834\\
\hline
Helicopter & 0.691&0.722&0.741&0.733&0.776&0.829\\
\hline
Rain&0.130&0.498&0.622&0.086&0.516&0.586\\
\hline
Rooster&0.827&0.883&0.967&0.838&0.926&0.957\\
\hline
Seawaves&0.514&0.612&0.693&0.563&0.662&0.714\\
\hline
Sneezing&0.683&0.828&0.916&0.693&0.888&0.960\\
\hline
MAP&0.609&0.745&0.795&0.615&0.788&0.822\\
\hline
\end{tabular}
}

\end{table}

One important advantage of weakly supervised AED using methods such as miSVM is that we can obtain a rough estimate of temporal location of events in an audio recording. In this section we evaluate different methods for temporal localization of events. We consider only audio events since audio events are usually short term acoustic phenomena.

Evaluating temporal localization is more challenging compared to recording level detection of events. Put more simply temporal localization evaluation implies instance level evaluation of all methods. Hence, we need strongly labeled \emph{test data}, so that the ground truth labels for instances (segments) are known. Since this is not available for the test data used in previous section, we use the left out fold from the ESC-10 dataset for evaluating instance-level performance. This means the fold which was left out of training process for simpleSWSL and graphSWSL is used as test data. Results accumulated over all $5$ runs are reported.   

Table \ref{tab:temploc} shows AP for instance-level detection of audio events. Once again we notice that SWSL based approaches gives $15-20\%$ absolute improvement over only weakly supervised learning. Graph based SWSL is once again superior to other methods. Overall this shows that our proposed method is suitable for temporal localization of audio events as well. 

\section{Conclusions}
     
We presented a novel learning framework named SWSL for learning simultaneously from strongly and weakly labeled data. Labeled data in these two different forms occur naturally for a variety of problems. The unified learning framework proposed in this paper allows one to leverage labeled data in both forms. More importantly, the proposed SWSL framework can help address concerns of only weakly supervised learning of audio events. Weakly supervised learning is a promising approach to scale audio event detection by exploiting weakly labeled web data. However, it comes with its own sets of concerns and we showed that SWSL can successfully address these concerns.  

From learning perspective, our primary assertion in this paper is that SWSL can be cast as a constraint form of semi-supervised learning. We proposed a method based on manifold regularization on graphs. This is in addition to the naive way of adopting strongly labeled data in weakly supervised learning. For both audio event and scene detection tasks a considerable improvement in performance can be observed by adding a small amount of strongly labeled data. This shows that a small amount of supervised data can mitigate the ill effects of signal and more importantly label noise in weakly labeled data obtained from web.    

Finally, our proposed SWSL is not restricted to the domain of audio content analysis and can be applied for other problems as well. Content based image retrieval and multimedia event detection where weakly supervised learning has been explored in depth are other suitable problem domains where SWSL can be applied.

\bibliographystyle{IEEEtran}
\bibliography{references}  



%

\end{document}